\documentclass{article}

\usepackage{arxiv}

\usepackage[utf8]{inputenc} % allow utf-8 input
\usepackage[T1]{fontenc}    % use 8-bit T1 fonts
\usepackage{hyperref}       % hyperlinks
\usepackage{url}            % simple URL typesetting
\usepackage{booktabs}       % professional-quality tables
\usepackage{amsfonts}       % blackboard math symbols
\usepackage{nicefrac}       % compact symbols for 1/2, etc.
\usepackage{microtype}      % microtypography
\usepackage{lipsum}

\usepackage[utf8]{inputenc}
\usepackage{booktabs} % For formal tables
\usepackage{subfig}
\usepackage{graphicx}
\usepackage{amsmath}
\usepackage[version=4]{mhchem}
\usepackage{siunitx}
\usepackage{longtable,tabularx}
\usepackage[color=green]{todonotes}
\usepackage{lettrine}
\usepackage{url}
\usepackage{algorithm}
\usepackage[noend]{algpseudocode}
\makeatletter
\def\BState{\State\hskip-\ALG@thistlm}
\makeatother
\usepackage{xspace}
\newcommand{\etal}{\emph{et~al.}\xspace}

\usepackage{verbatim}
\usepackage{textcomp}
\usepackage{array,multirow}
\usepackage{makecell}
\usepackage[toc,page]{appendix}

\title{A Test-bed For Measuring UAS Servo Reliability}

\author{AbdElRahman ElSaid \\
Computer Science \\
Rochester Institute of Technology\\
Rochester, New York, 14623 \\
\texttt{aelsaid@mail.rit.edu}
\And
Daniel Adjekum \\
Department of Aviation \\
University of North Dakota \\
Grand Forks, North Dakota 58202 \\
\texttt{daniel.adjekum@und.edu}
\And
John Nordlie \\
Computer Science Department \\
College of Engineering \& Mines \\
University of North Dakota \\
Grand Forks, North Dakota 58202 \\
\texttt{john.nordlie@und.edu}
\And
Fatima El Jamiy \\
Computer Science Department \\
College of Engineering \& Mines \\
University of North Dakota \\
Grand Forks, North Dakota 58202 \\
\texttt{fatima.eljamiy@und.edu}
}

\begin{document}
\maketitle

\begin{abstract}
		Extant literature suggests minimal research in the area of system reliability of components used in the design of these UAS (Unmanned Air Systems), thus, subjecting UAS to critical failures that may pose a safety hazard to flight operations.
		The purpose of the study was to critically assess the reliability of a laboratory designed UAS component test-bed operated using real-world data collected from a Boeing Scan Eagle\textsuperscript{\textregistered} UAS aileron servo unit via a flight data recorder. The study hypothesized that the test-bed's unit replicating a UAS aileron servo motor's reliability, in terms of a base-line measured encoder output of commanded servo positions, will not be significantly different after double and triple periods of time for continuous operating cycles.		
		Study adds to paucity of extant research on UAS reliability and recommends further studies on commercial UAS components reliability and time to failure.
\end{abstract}

\section{Introduction}
\lettrine{T}{he} seemingly rapid evolution of Unmanned Aerial Systems in contemporary times have shifted the operational paradigms in aviation and has also brought contemporary challenges related to the development of safer and efficient UAS within National Airspaces (NAS) globally. Despite the obvious proliferation in the use of UAS in various operational activities both civil and military, one of the numerous challenges faced in the design and fabrication of UAS has been a means of assessing the reliability of systems components over varying operational cycles to determine critical failure characteristics with time. Of most concerns has been the proliferation of commercials-off-the-shelf (COTS) components used in the building of small UAS (55 lbs. and below), used mostly for recreational and civil-related activities~\cite{rodriguez2017small, jang2007small, hague2006field, murch2011low, bernard2017using, logan2005technology, rangel2009development, morris2004examples}. \\
% \lettrine{T}{he} era of unmanned aviation is flourishing and advancing in leaps and bounds jumping over all challenges and obstacles reflect a natural expansion of this promising technology.
According to the FAA\footnote{Federal Aviation Administration} 2019-2039 Aerospace Forecasts for UAS\footnote{Unmanned Air Systems (also known as Unmanned Air Vehicle UAV)}, the non-model UAS (non-hobby) registration reached 277,000 in the fourth quarter of 2018, while the model UAS (hobby) registration reached 903,588 in the same quarter. Wargo~\etal~\cite{wargo2014unmanned} forecasts the UAS future to have a market annual demand of about 70,000 aircraft for state demand, and 170,000 aircraft for the commercial sector, by the year 2035.
One of the pillars of the success of UAS technology is the availability and cost-effectiveness of its components or sub-systems; Most UAS systems and components are within the reach of any interested organization or individual.
This means that most of these subsystems are COTS that are ready to use after implementation, which makes it convenient to design and construct a fully functioning UAS that meets many real-life applications.

However, COTS convenience often comes with a a reliability price: they are usually sold with no-liability or limited-liability to its manufacturers. This serious problem plagues the fledgling UAS original equipment manufacturing (OEM) industry and potentially affects progress towards technological and economic maturity for these stakeholders and jeopardizes its path to maturity. Unlike manned aviation systems, UAS's sub-systems lack the exhaustive reliability tests in general and destructive tests in particular~\cite{freeman2014actuation}. 

Logan~\etal~\cite{logan2005technology} at the NASA Langley Research Center conducted a study about `technology challenges in small UAV development'. After conducting tests over 30 combinations of different vendor UAV COTS components, including servos, they found that `over 60\% of the configurations tested had some form of problem consisting of either erratic behavior, jitter, control reset, overheating, or system instability over time'. The study concluded that `improvements in avionics reliability, stability, and compatibility are a clear need, particularly in low-cost applications.' 
Therefore, in most cases, there are concerns regarding the safety and operational capabilities of UAS to perform BVLOS\footnote{Beyond Visual Line Of Sight} missions, where the aircraft is controlled, either autonomously, or remotely from no-visual-contact distance.

As part of a contemporary push for more data-driven research and improvements in operational capabilities of UAS, a collaborative effort by the University of North Dakota and the State of North Dakota yielded a certificate of authorization (COA) for the conduct of BVLOS flights by UAS within a specified corridor of airspace in North Dakota\footnote{\url{https://www.unmannedsystemstechnology.com/2019/01/faa-grants-first-national-waiver-for-uas-operations-over-people-and-bvlos-flights/}}. This COA may have been granted by the FAA due to the high fidelity of data collected over numerous research and suggest that with increasing reliability of components used in UAS fabrications such feats are possible and regulators can give exemptions once performance-based navigational criteria are met.

% As a good example of the important role of research and data analysis in the realm of UAS, The University of North and the State of North Dakota has recently been given a certificate of authorization to fly BVLOS within the region due to high fidelity of data collected over numerous research and that suggest that with increasing reliability of components used in UAS fabrications such feats are possible and regulators can give exemptions once performance based navigational criteria are met. \footnote{\url{https://www.unmannedsystemstechnology.com/2019/01/faa-grants-first-national-waiver-for-uas-operations-over-people-and-bvlos-flights/}}. Given that the population of the state of North Dakota is not as dense as in other states and regions, such studies and analysis should be more emphasized.

One of the essential components to fixed-wing UAS flight-control, to maintain a safe flight, is the device which actuates its control surfaces. Electro servo motors are the most convenient choice to act as this actuating system in small UAS in that it is relatively cheap, compact in size, precise, controllable, and light-weight~\cite{ohanian2012piezoelectric}. Though servos have all these advantages, they still dominantly suffer from limited information about reliability studies and examinations for reasons which relate to commercial competition and/or cost. Accordingly, this study focuses on the design and implementation of a destructive testing platform to measure the time to failure (TTF)\footnote{also known as operational life expectancy} of electric servos of types which are typically used in UAS control. 

In terms of the method for this research, a case-study approach is used to obtain the data of an operating servo for the study and then a quantitative analysis is conducted on the outputs of  
operational servo's real-time actuating angles to determine the 
% a servo actual operational angles to determine
frequencies of the controlled surface movements. After that, the force conditions that servos normally operates in are calculated to simulate those on the proposed test-bed which will be designed. Finally, a test-bed and its control system are actually designed and built for the experimental portion of the study, which aims at collecting servo's commanded position data and actual measured servo movements (using a digital optical rotation encoder) through various time periods. It will also simulate the forces acting on the servo in real-time. 

It is hypothesized that with high reliability, system components such as the servo (using test-bed) operating under varying simulated flight conditions will not have significant differences in the mean outputs of an encoder recorded position for similar commanded inputs (angles) over the simulated period of time.
% This work begins with a case study to pick a servo. Then a statistical analysis is performed on the servo's actual operation-angles to determine the frequency of the controlled control-surface movements. After that, the force conditions that the servo normally operates in are calculated to simulate those forces on the testing-bed. Last but not least, the test-bed and its control system were actually built for the study's experiments to command the servo angles, collect the servo's position data, and simulate the forces acting on the servo in real-time operations.

%
%
% you need some citations on the current usage of UAS globally or within the US at least. You also need to provide some statistics on usage and the type of use. You also have to specify the types ( Large, medium, small etc) to inform the reader about the scale of proliferation within the aviation environment.
%
%  Example:  FAA (201X) suggest that there are currently about 20 million registered small UAS ( less than 55 pounds) in the US and over 200,000 medium to large UAS ( over 55 pounds ) in the US National Airspace (NAS). Boeing ( 2018) predicts that the number of UAS in the NAS will be around 40 million in the year 202X ( Boeing, 2019).
%
% These hypothetical citations will set a very good background and context for the reader to understand the scale and complexity of UAS operations and the need for it to be safe in the NAS.
\section{Related Work}
\label{sec:related_work}

Reliability study is essential in the aviation world. However, 
empirical studies related to reliability of UAS components and sub-components have been mostly limited as compared to manned aircraft~\cite{dermentzoudis2004establishment, neogi2007engineering}. With the recent proliferation in the operations of UAS and attendant safety concerns raised by regulators and researchers, it is important for a much broader focus to be placed on system reliability especially in the commercial-off-the shelf UAS market. 
% the UAS studies are still not well established. Yet, it is gaining potential because of the safety concerns raised by regulators and feasibility researchers.

Casewell~\&~Dodd~\cite{caswell2014improving} suggested that that 25\% of the UAS failures are attributed to electronic components failures, and the precent is subject to increase as control software improves and less redundancy is used in the electronic components.
Their study also mentions that the commercial-grade UAS components have a failure rate two to three times more than their military-grade counter parts. The commercial components reliability the study mentioned relies on Telcordia SR332, which is a standard use as a reference based on the components manufacturing entities.

Dermentzoudis~\cite{dermentzoudis2004establishment} established methods to collect data to be used in evaluating UAS reliability. The study is based on data collected from military aircraft based on mass data collection from large fleets. 
% The study focus on holistic methods rather than individual sub-systems.
It also relies on the mass collection of data of relatively large fleets, which is not necessarily available for commercial application. This might indicate that the study is of a holistic nature, which is built on statistical concerns about the whole UAS system rather than individual sub-systems and how long they can serve soundly.

Uhlig~\etal~\cite{uhlig2006safety} studied COTS as a cause of failure in UAS systems. The study investigated the integration of the specific sub-systems and added some redundancy to the components used to see its influence on the reliability of the UAS. The study does not offer a statistical measures.

Bhamidipati~\etal~\cite{neogi2007engineering} introduced an important study about COTS used in UASs and their reliability. 
% The study focused on the loss of power due to lack of information, or incorrect estimate, of available fuel quantity.
Though the study is intensive, the experiments were done on a single aircraft and its engine, fuel throttling, and burning system. The study can be much advanced if data were collected from several aircraft and engines of the same model to enrich the results.

Petritoli~\etal~\cite{article22, petritoli2018reliability} derived a model that can be used to perform an overall evaluation of the reliability of the UAS systems to improve the logistics of UAS maintenance plans. The work is very well represented but still there was no mention to a reliable tool to collect the data necessary to evaluate the systems' reliability and availability. 
% The study aimed to determine the appropriate preventive maintenance intervals for the UAS systems.

Tan~\etal~\cite{tan2017research}, as part of a study, suggested a reliability model for UAS that considers
the human factor. The study shows important results but was not much concerned with the hardware components of UASs and how their unreliability can much affect the human operator performance.

Pan~\cite{pan2017hybrid} investigated the use of hybrid data collection to detect anomalies which influence UAS reliability. The data collected was general performance data for the aircraft systems like throttle, thrust, autopilot mode, altitude, static pressure, and other flight parameters.
The mixture of data along with data from command switches were exploited to filter out unusual behavior in the UAS system as a whole. The study's contribution is an anomaly detection model rather than a direct reliability study for the UAS systems.

It is well known that reliability of aircraft equipment is very important to aircraft during the flight because performance of aircraft product can affect flight safe directly. There are many methods available to assess the reliability of components and sub-system components. According to the FAA Handbook on Safety Risk Analysis\footnote{Federal Aviation Authority Hand Book on Risk Management. \url{https://www.faa.gov/regulations_policies/handbooks_manuals/aviation/risk_management/ss_handbook/media/Chap9_1200.pdf}} One of the effective methods is the  Failure Mode Effect and Criticality Analysis (hereafter called FMECA), which is extensively used in reliability analysis and valuation . FMECA is designed to analyze all sorts of the potential failure in each component, and by analyzing and computing criticality, FMECA may tell the incoming failure and its effect, that is in order to make the equipment work normally, FMECA is applied in an aircraft equipment to analyze its reliability and improve operational reliability of the product, especially over a period of time.

LiJun and Xu~\cite{jun2012reliability} conducted a study on aviation components reliability using mathematical models and the average of operational time was predicted based on calculating failure probability of all electrical components. According to the process of reliability theory FMECA, all kinds of the failure mode, reasons, effects and criticality of the products could be determined completely. The results indicate that application of FMECA method can analyze reliability in detail and improve operational reliability of the equipment. Their study provided theoretical bases and concrete measures of maintenance of the products to improve operational reliability of products.

According to Moubray ( 1997), there are basically six unique failure patterns of components and equipment in aviation~\cite{poddar2015reliability}.  The ``bathtub curve'' Failure Pattern has a high probability of failure when the equipment is new, followed by a low level of random failures, and followed by a sharp increase in failures at the end of its life. This pattern accounts for approximately 4\% of failures. The ``wear out curve'' failure pattern consists of a low level of random failures, followed by a sharp increase in failures at the end of its life. The pattern accounts for approximately 2\% of failures.

The ``fatigue curve'' is characterized by a gradually increasing level of failures over the course of the equipment’s life. This pattern accounts for approximately 5\% of failures. The ``initial break in curve'' starts off with a very low level of failure followed by a sharp rise to a constant level. This pattern accounts for approximately 7\% of failures. The “random pattern” failure is a consistent level of random failures over the life of the equipment with no pronounced increases or decreased related to the life of the equipment.  This pattern accounts for approximately 11\% of failures. Finally, the ``the infant mortality curve'' shows a high initial failure rate followed by a random level of failures. This pattern accounts for 68\% of failures.

The investigation of Long~\etal~\cite{long2007empirical} provides considerable insight regarding potential impact of MTBF\footnote{Mean time between failures} analysis on reliability and reduction in direct operational cost for UAS operations. Using the Predator and Global Hawk platforms, they assessed increases in MTBF over time as a metric for reliability growth and improvement. The study suggested that increasing MTBF was generally associated with increasing reliability, if failures are exponentially distributed — that is, if failures are random and do not impact one another. 

Interestingly, Long~\etal also carefully untangled all of the program costs and were able to determine the dollars devoted just to improvements in reliability over the time. They found that for Global Hawk, system failure rate was reduced by 42\% from 2001 to 2006 and life cycle support costs were reduced by 23\%.  This finding allowed them to calculate a return on reliability dollar investment (RORI) of 5:1.  For Predator, the figures were even more impressive.  Between 1998 and 2006, improvements in platform reliability reduced failure rates by more than 48\% and reduced life cycle support costs by 61\%.  The RORI in this case was 23:1. 

\section{Methods and Materials Section}

\subsection{Case Study}
\label{sec:case_study}

The objective of this part of the study is to investigate the effectiveness of the introduced test-bed using data obtained from the deployments of the servo of 
The Boeing Insitu ScanEagle\textsuperscript{\textregistered}, which is recorded on the aircraft's FDR\footnote{Flight Data Recorder}. This aircraft is a small (3.9 feet long, 10.2 feet wingspan, 39.7 lbs) and long-endurance (20+ hours) UAS that can fly in BVLOS operations with a speed of 55-80 mph and a ceiling altitude of 16,000 feet. The aircraft is equipped by a 3W\footnote{\url{http://3w-international.com}} 2-stroke piston engine; 1.5 horsepower engine\footnote{\url{https://www.af.mil/About-Us/Fact-Sheets/Display/Article/104532/scan-eagle}}.

A pre-statistical analysis was performed on the data to define the ailerons' most frequent positions as well as outliers in these positions, which are possible errors in read signals. The 7854 entires obtained from the FDR file were used to calculate the frequencies of the servo commanded positions. These frequencies were then normalized and used to repeat these positions in the experiments in proportional to the frequencies. The most extreme outliers were considered as bad recorded signals when they are not realistically possible (for instance when the deployment angle is greater than $90^o$ or less than $-90^o$).

\subsection{Forces Calculation}
Using the specification of the aircraft's aileron shown in Table~\ref{tab:ac_specs}, the forces exerted on the control surface by aerodynamics was calculated. Taking the air density to be the standard sea-level density: $\rho=1.27 kg/m^3$, the forces were calculated using the equation: $F = \frac{1}{2} \rho v^2 \times COSINE(\theta)$ where $F$ is the normal force on the control surface, $\rho$ is the air density, $v$ is the aircraft speed, and $\theta$ is the angle of the control surface to the horizontal.

\begin{table}[h]
	% \small
  \centering
  \caption{Aircraft Specifications}
  \label{tab:ac_specs}
  \begin{tabular}{llllll}
    \toprule
    \textbf{Aileron Width}  						& $0.428  m$		\\
	\textbf{Aileron Height}							& $0.048  m$		\\
	\textbf{Aileron Surface Area} 					& $0.021  m^2$		\\
	\textbf{Max. Cruise Speed}						& $50  m/h$			\\
	\textbf{Min. Cruise Speed}						& $40  m/h$			\\
	\textbf{Arg. Cruise Speed}						& $45  m/h$			\\
	\textbf{Max. Aileron Deployment Angle (cruise)} & $40^\circ$		\\
    \bottomrule
  \end{tabular}
\end{table}

% \begin{equation}
% 	\label{eq:force}
% 	F = \frac{1}{2} \rho v^2 \times COS(\theta)
% \end{equation}

% where:
% $F$: The normal force on the control surface,
% $\rho$: The air density,
% $v$: The aircraft speed, and
% $\theta$: The angle of the control surface to the horizontal.

\subsection{Forces Simulation}
\label{subsec:forces_calc}
\begin{figure}
	\centering
\includegraphics[width=.98\textwidth]{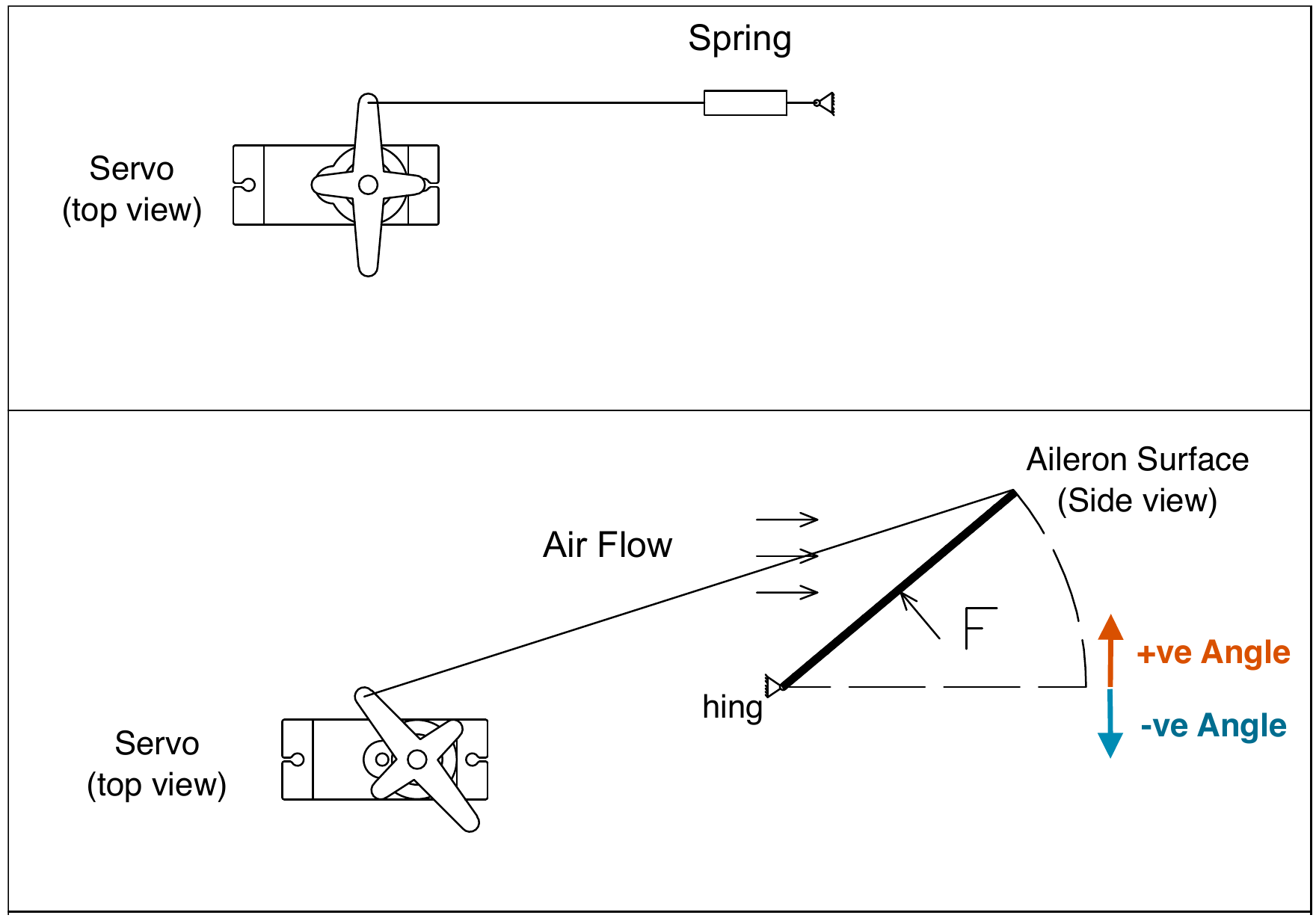}
\caption{\label{fig:surfaceServoSys}Servo/Control-Surface System (self-designed).}
\end{figure}

To simulate the force acting on the control-surface, and consequently on the servo, an extension constant force mechanical spring (force relates linearly to spring extension) was used (Figure~\ref{fig:surfaceServoSys}). The force which acts on the control-surface changes as the angle of the control-surface changes. The spring was picked so that its resistance is linear and obeys Hooks law: $F=kx$, where $F$: spring's resistance, $k$: spring's stiffness, and $x$: is the length of the extension of the spring. The maximum force acting on the control surface (at the maximum deploying angle) should be covered in the maximum extension of the spring. 

\begin{table}[h!]
\centering
  \caption{Negative Positions Statistical Description}
  \label{tab:neg_stat}
  \begin{tabular}{llr}
    \toprule
    \textbf{N} 						& Valid 	& 1095		\\
      								& Missing 	& 0			\\
	\textbf{Mean}\hspace{1cm}					&			\\
	\textbf{Std. Error of Mean} 	&			& .28664	\\
	\textbf{Median}					&			& -11.2503	\\
	\textbf{Std. Deviation}			&			& 9.48505	\\
	\textbf{Variance}				&			& 89.966	\\
	\textbf{Skewness}				&			& -2.701	\\
	\textbf{Std. Error of Skewness} &			& .074		\\
	\textbf{Kurtosis}				&			& 10.336	\\
	\textbf{Std. Error of Kurtosis}	&			& .148		\\
	\textbf{Range}					& 			& 73.13		\\
	\textbf{Minimum}				&			& -74.53	\\
	\textbf{Maximum}				&			& -1.41		\\
	\textbf{Percentiles} 			& 25		& -11.2503	\\
									& 50		& -11.2503	\\
									& 75		& -5.6252	\\
    \bottomrule
  \end{tabular}
\end{table}

\begin{table}[h!]
	\centering
  \caption{Positive Positions Statistical Description}
  \label{tab:pos_stat}
  \begin{tabular}{llr}
    \toprule
	\textbf{N} 								& Valid		& 6557		\\
											& Missing	& 0			\\
	\textbf{Mean}							&			& 16.2681	\\
	\textbf{Std. Error of Mean}				&			& .13087	\\
	\textbf{Median}							& 			& 14.0629	\\
	\textbf{Std. Deviation}					&			& 10.59699	\\
	\textbf{Variance}						&			& 112.296	\\
	\textbf{Skewness}						& 			& 1.912 	\\
	\textbf{Std. Error of Skewness}			&			& .030		\\
	\textbf{Kurtosis}						&			& 5.957		\\
	\textbf{Std. Error of Kurtosis}			&			& .060		\\
	\textbf{Range}							&			& 87.19		\\
	\textbf{Minimum}						&			& 1.41		\\
	\textbf{Maximum}						& 			& 88.60		\\
	\textbf{Percentiles} 					& 25		& 9.8441	\\
											& 50		& 14.0629	\\
											& 75		& 19.6881	\\
    \bottomrule
  \end{tabular}
\end{table}

\subsection{Data Preparation}
Boeing Insitu ScanEagle\textsuperscript{\textregistered} is equipped with a Horizon MicroPilot\textsuperscript{\textregistered}\footnote{\url{https://www.micropilot.com}} flight controller. Flight data was downloaded after each flight using a MicroPilot software product called Logviewer\textsuperscript{\textregistered} using a computer.

The study's used data file is 7854 entires representing data points collected from a point-to-point flight. The aircraft's autopilot samples data at the rate of 5 samples per seconds. This means that the data covers about 26.2 mins flight. The maximum altitude reached in the recorded flight was 2267 feet. The FDR saves the aileron servo position in measurement units called 'fine servo units'. These units range between -32,767 and +32,767 and had to be converted to rotational degrees given that the servo rotation range is $[0\textsuperscript{\textdegree}, 180\textsuperscript{\textdegree}]$.

An FDR log file was used to statistically study the positions of the servo during flight. 
% The results of this study is shown in Table~\ref{tab:neg_stat} for the negative aileron angles, and Table~\ref{tab:pos_stat} for the positive aileron angles. 
% Figures~\ref{fig:histo_pos_postions}, and~\ref{fig:histo_all_postions}, show the frequencies of the angles in the FDR file.
% Box-plot shown in Figure~\ref{fig:box_plot} show the confidence interval of the positions in the FDR file.
% The outliers are also shown in the figure.
% The 50\% confidence interval shows that maximum and minimum positions are $9.84^\circ$, and $19.69^\circ$ respectively, while for the negative positions are $-11.25^\circ$ and $-5.63^\circ$ respectively.]
% The data had a mean of $16.27^o$ and a standard deviation of $10.597$ ($N= 6,557$) for the positive positions, and a mean of $-11.60^o$ and a standard deviation of $9.485$ ($N=1,095$)for the negative positions.
% , based on the data median in both data groups (refer to Figure~\ref{fig:surfaceServoSys} for angle sign).

The negative positions ($N = 1095$) had a mean angular displacement of -11.60 degrees and standard deviation of 9.485. The positive positions ($N = 6557$) had a mean angular displacement of 16.27 degrees and standard deviation of 10.597. Table~\ref{tab:neg_stat} and Table~\ref{tab:pos_stat} highlights the details of the descriptive statistics respectfully. As depicted, the number of positive aileron positions exceed the number of negative positions, which is expected because negative position acts as an aid for the positive position on the opposite side aileron in severe maneuvers~\cite{anderson2010fundamentals, anderson1999history}.

A list of positions ( positive and negative) was prepared based on FDR data output to reflect the angles that the servo operate in real life and their frequencies. This list took into consideration the real-life servo operation and represented this in the number of times each position appear in the list. Table~\ref{tab:position} depicts a sample of the prepared list based on the FDR file.

\begin{table}
	\centering
% \fontsize{8pt}{5pt}\selectfont
  \caption{Commanded Positions}
  \label{tab:position}
  \begin{tabular}{cccccccccccccccccccccccc}
    \toprule

4	& -11	& 13	& -11 	& 32	& 15	& 7		& 17	& 23	& 14	%\\
3	& 11	& 11	& 11	& 10	& 17	& -1	& -11	& 18	& 6		\\
25	& -10	& 25	& 20	& 11	& 10	& 7		& -24	& 7		& 37	%\\
8	& 14	& -18	& -14	& 10	& 11	& -4	& 17	& 35	& 6		\\
3	& 14	& 28	& 17	& 34	& -6	& 11	& -30	& 4		& 13	%\\
17	& 21	& 11	& 15	& -20	& -1	& 24	& 20	& 6		& -11	\\
14	& 13	& 24	& -25	& 21	& 8		& -13	& 23	& 14	& -15	%\\
3	& 14	& 17	& 30	& 31	& 15	& -28	& 14	& -11	& 15	\\
1	& 27	& 13	& -8	& 8		& -3	& -7	& 10	& 8		& 10	%\\
7	& 18	& 27	& 14	& 20	& 8		& 4		& 13	& 15	& -17	\\
23	& -11	& -3	& 18	& 8		& 20	& 13	& -21	& 6		& -27	%\\
-23	& 21	& 15	& 13	& 10	& 13	& 28	& 15	& 7		&		\\
    \bottomrule
  \end{tabular}
\end{table}

% \begin{figure}
% % \begin{sidewaysfigure}
% \centering
% \subfloat[Training error for 1 second predictions.]{\includegraphics[width=.48\textwidth, height=.20\textheight]{./img/{cost_fun_all_art_1}.pdf}
% \label{fig:art_all_cost_1sec}}
% % \quad
% % \\
% \subfloat[Training error for 5 second predictions.]{\includegraphics[width=.48\textwidth, height=.20\textheight]{./img/{cost_fun_all_art_5}.pdf}
% \label{fig:art_all_cost_5sec}}
% \\
% \subfloat[Training error for 10 second predictions.]{\includegraphics[width=.48\textwidth, height=.20\textheight]{./img/{cost_fun_all_art_10}.pdf}
% \label{fig:art_all_cost_10sec}}
% % \quad
% % \\
% \subfloat[Training error for 20 second predictions.]{\includegraphics[width=.48\textwidth, height=.20\textheight]{./img/{cost_fun_all_art_20}.pdf}
% \label{fig:art_all_cost_20sec}}
% \caption{Mean squared error during the training process for the three architectures predicting vibration in 1, 5, 10, and 20 seconds in the future.}
% \label{fig:art_all_costs}
% % \end{sidewaysfigure}
% \end{figure}
% \input{04-dataPreparation}
\section{Implementation}
The main concept with the test-bed was to let the servo operate in all the positions that were collected from the statistical study performed on the FDR file. 
The design of the test-bed platform exploited a digital rotary encoder to log the output position of the servo when it reached its commanded input position. 
The operations considered the frequency of the position such that the servo on the test bed runs more on the positions that occurred in the real-life operations, to get a more realistic result. The servo kept operating and its actual positions were recorded along with their corresponding commanded positions, meaning the inputs and outputs of the servo were logged. The logged data were compared to determine if there were any significant differences in the mean values of the input and output of servo displacement over the simulated period (time to failure).
% the time the servo took to malfunction. 

% As mentioned previously, the force on the servo will be simulated to achieve realistic results.
% The design of the platform exploits a digital rotary encoder to record the position of the servo when it reaches its commanded position.

\subsection{Hardware}
\subsubsection{CAD Design}
After calculating the forces acting on the servo and choosing the spring to simulate these forces, a CAD\footnote{Computer Aided Design} was used to build a 3-D design for the platform. One of the major points which were considered in this step of the study was the alignment of the servo shaft and the encoder shaft. This is particularly important because any misalignment between the two shafts would result in additional forces, other than the ones actually acting on the servo and considered in the design. 

\begin{figure}
\centering
\subfloat[\normalsize 3-D Wire View.]{\includegraphics[width=.48\textwidth]{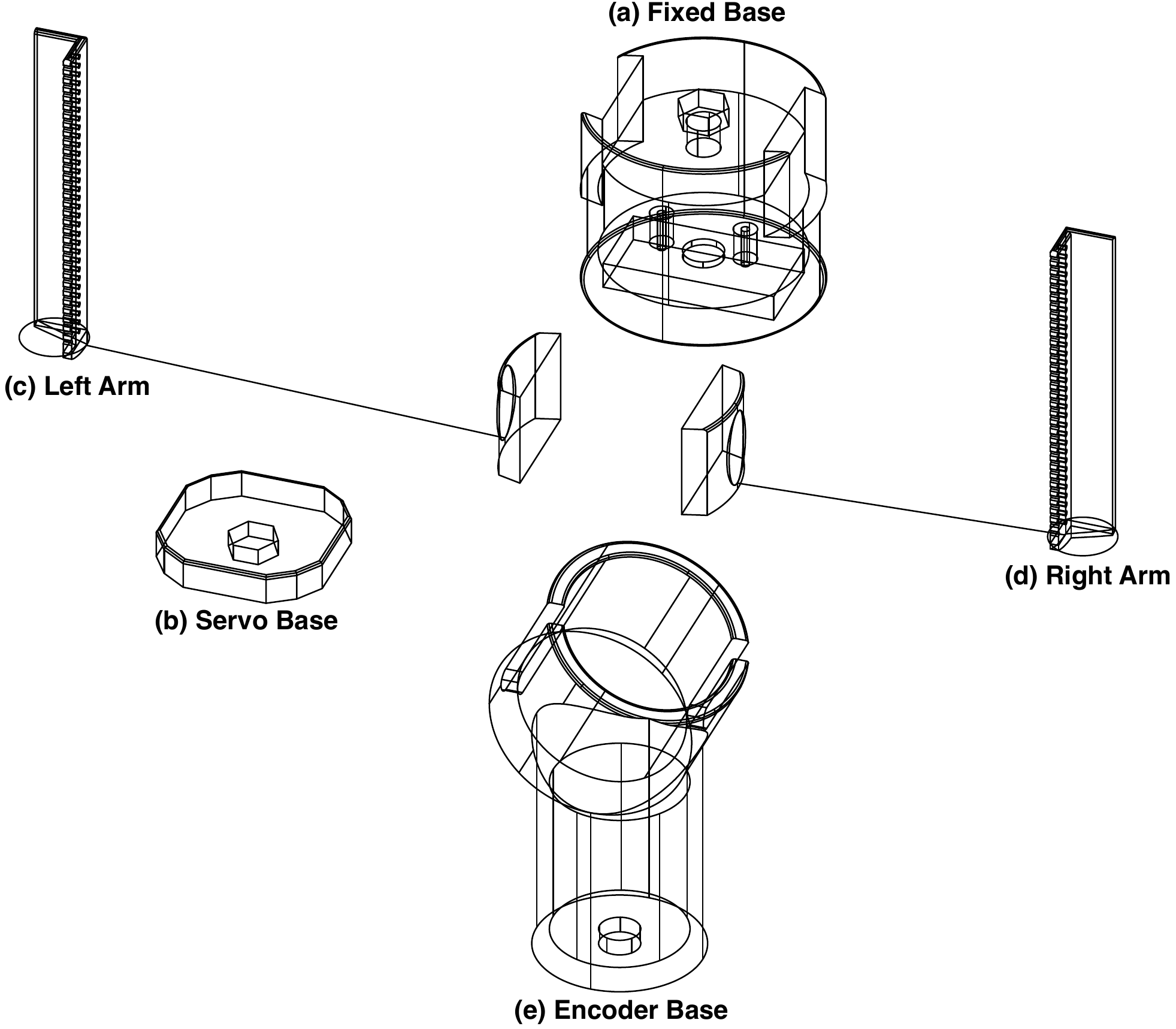}
\label{fig:3dwire}}
% \\
\subfloat[\normalsize 3-D Realistic View.]{\includegraphics[width=.48\textwidth]{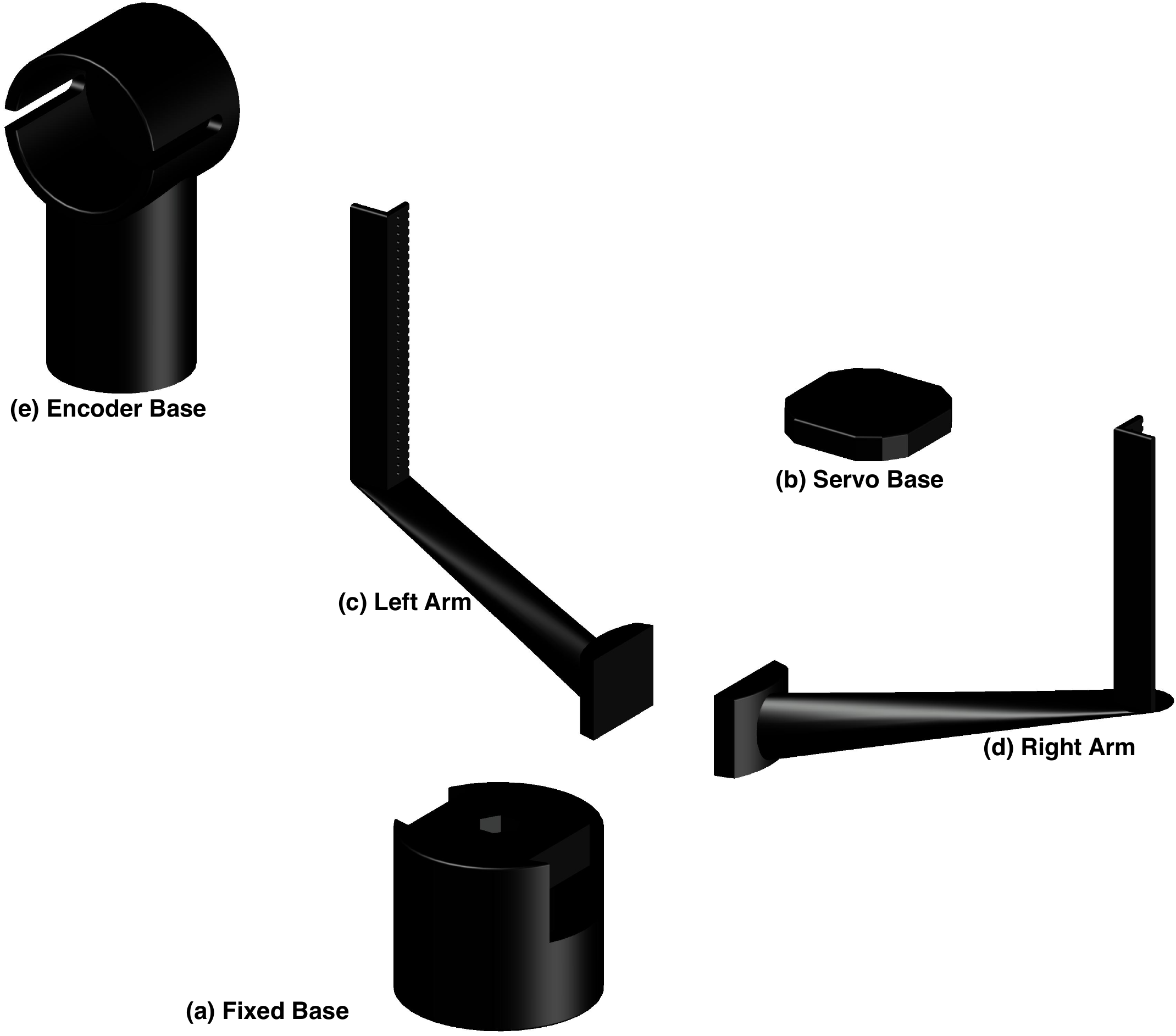}
\label{fig:3dreal}}
\caption{3-D Design (Design implemented as part of the study)}
\label{fig:3ddisgn}
\end{figure}

Figure~\ref{fig:3dwire} shows the CAD 3-D drawing for the platform and its internal composition. Figure~\ref{fig:3dreal} shows the CAD 3-D a realistic drawing for the platform.

\begin{figure}
\centering
\subfloat[\normalsize Disassembled Parts.]{\includegraphics[width=0.48\textwidth]{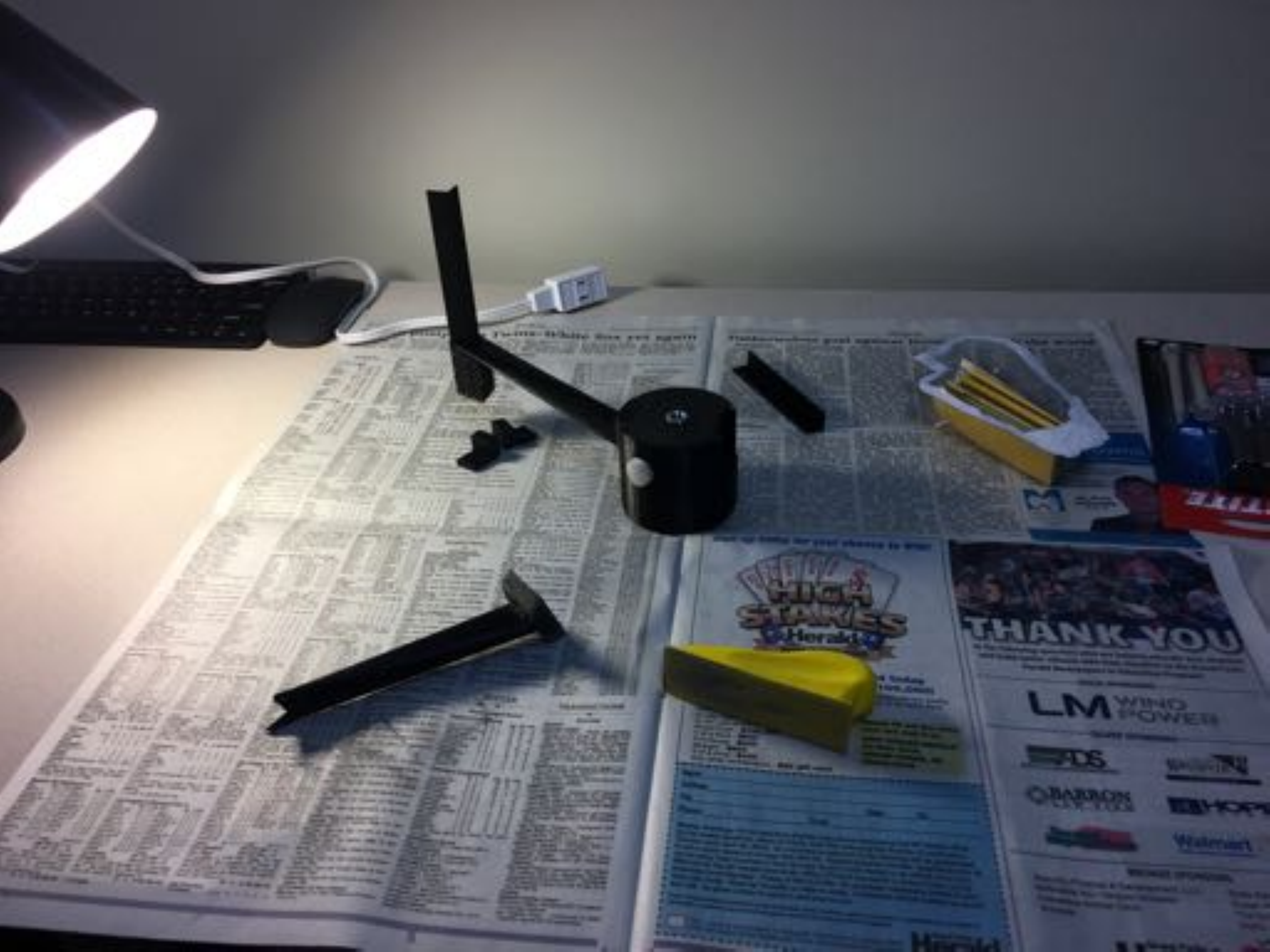}
\label{fig:parts1}}
\subfloat[\normalsize Servo Base's Magnet]{\includegraphics[width=0.48\textwidth]{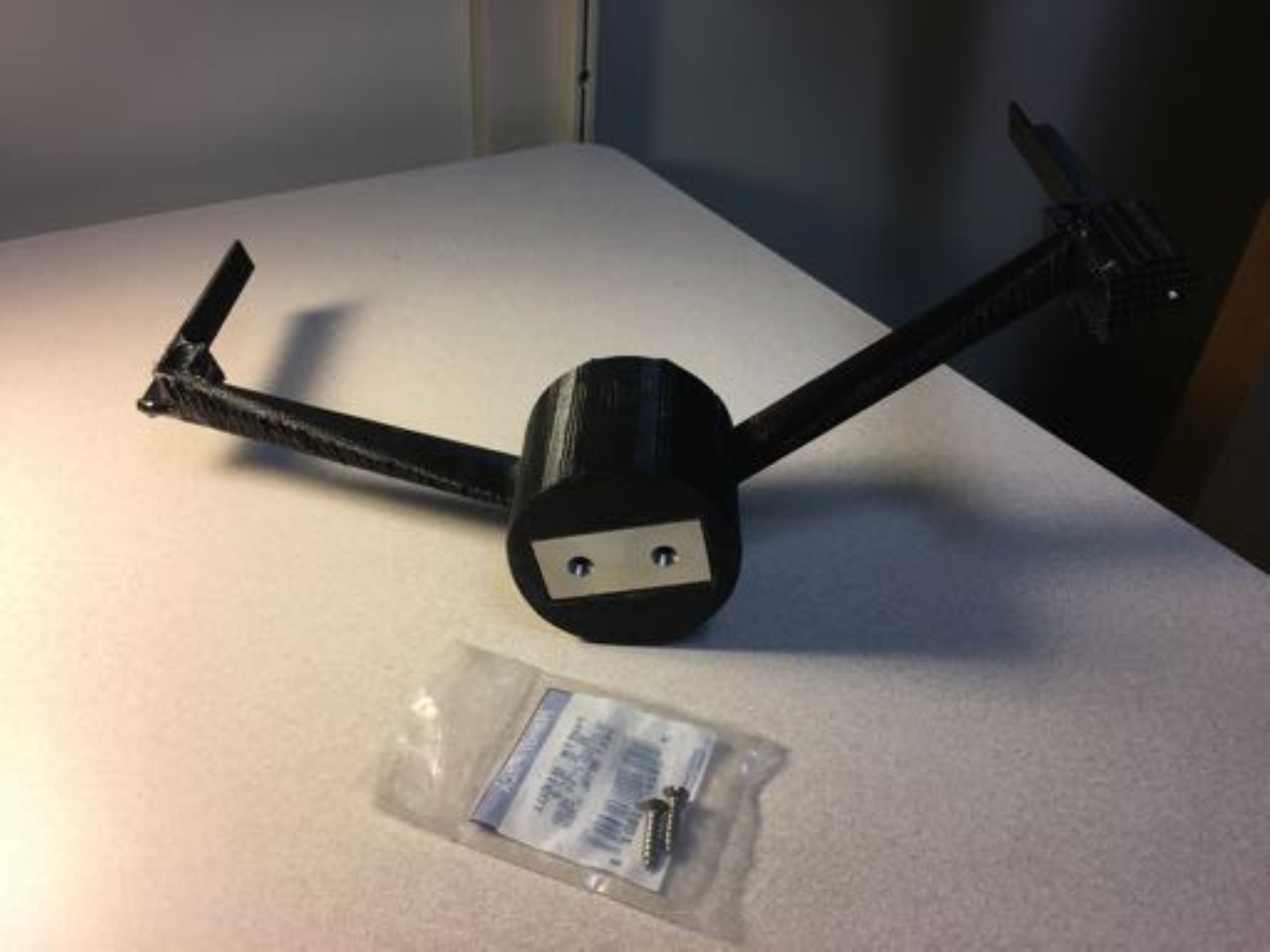}
\label{fig:parts2}}
\\
\subfloat[\normalsize Servo Base and Encoder Base.]{\includegraphics[width=0.48\textwidth]{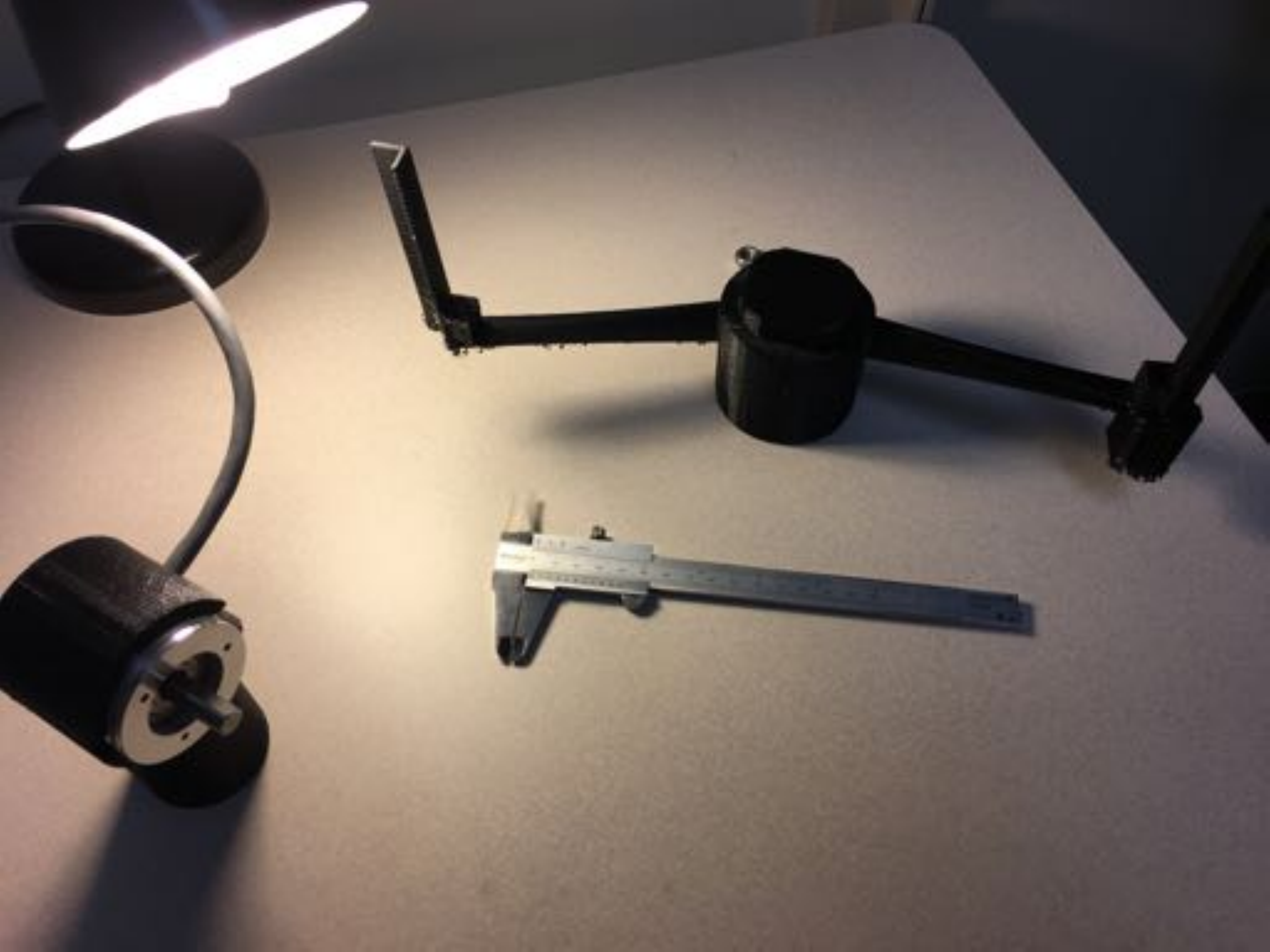}
\label{fig:parts3}}
\subfloat[\normalsize Assembled Platform..]{\includegraphics[width=0.48\textwidth]{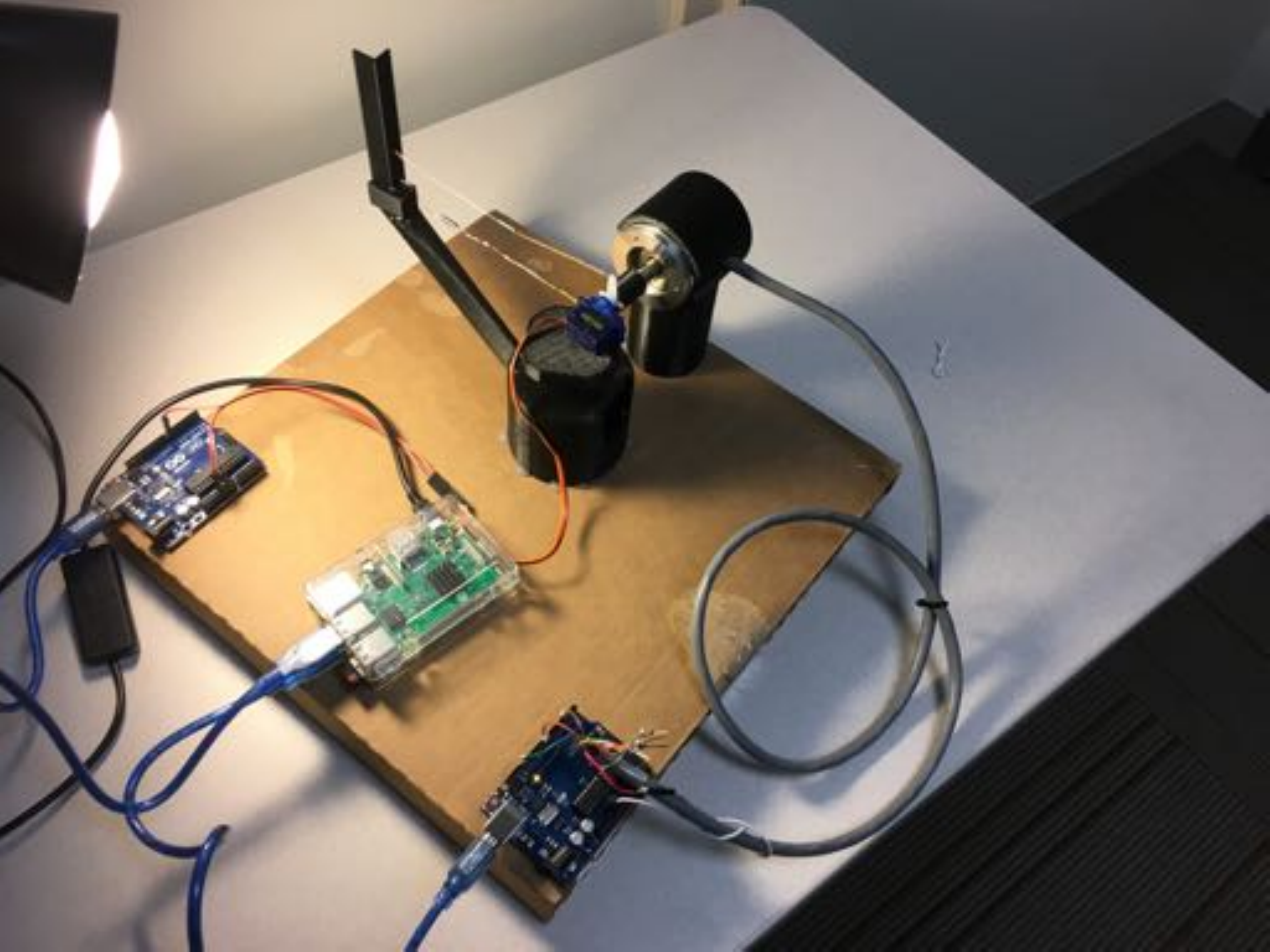}
\label{fig:assem}}

\caption{3-D Printed Parts (Parts printed as part of the study).}
\label{fig:parts}
\end{figure}

The platform consisted of a 
% \textbf{\it a)}
Fixed-base for the servo which can move on the X-Y horizontal plane (Figure~\ref{fig:parts1}). 
A magnet was fixed at the bottom of the base to let it move in the X-Y plane on a gallivanted steel sheet. This 2-D movement allows for adjustment of the servo position to the encoder position. 
The fixed base has a moving base above it where the servo is attached. 
% \textbf{\it b)}
The Servo-base is allowed to move in the Z-direction to further adjust the servo position with respect to the encoder position (Figure~\ref{fig:parts2}). The Servo-base is fixed to the Fixed-base by a bolt which screws into the nut in fixed into the Fixed-base.
After the servo in the right position, the Servo-base is tightened using another nut to restrict it from moving up and down during the test. 
The 
% \textbf{\it c)}
Right-arm and the 
% \textbf{\it d)}
Left-arm (Figure~\ref{fig:parts3}) are used to hinge the load springs from one side.
The \textbf{\it e)} Encoder-base houses the optical-digital encoder and it purely fixed to the ground (Figure~\ref{fig:assem}).

Figures~\ref{fig:parts} show the 3-D printed parts of the platform and Figure~\ref{fig:assem} shows the assembled testing-bed~\footnote{A sample video for the operating test-bed is at: \url{https://www.youtube.com/watch?v=OtsnsTmVvNw}}.

% \begin{figure}
% \includegraphics[width=.98\textwidth, height=.35\textheight]{./img/assem.pdf}
% \caption{\label{fig:assem}Assembled Platform.}
% \end{figure}

\subsubsection{Microcontrollers and Computer}
The study used two Arduino UNO\textsuperscript{\textregistered} controlling chips. Each chip has a ATmega328 microcontroller with 32 Kb flash memory, 2 kb SRAM and a 20 MHz oscillator. The first one was used to receive input commanded servo positions (one at a time) from a computer and command the servo position by sending to the servo one position at a time. 
The second microcontroller was used to read the position output by the rotary encoder and send it back to the computer for logging.
The two controllers have a 16 Mhz clock speed. Serial ports were used to communicate with the controllers. 

\begin{figure}[h]
	\centering
\includegraphics[width=.70\textwidth, height=.13\textheight]{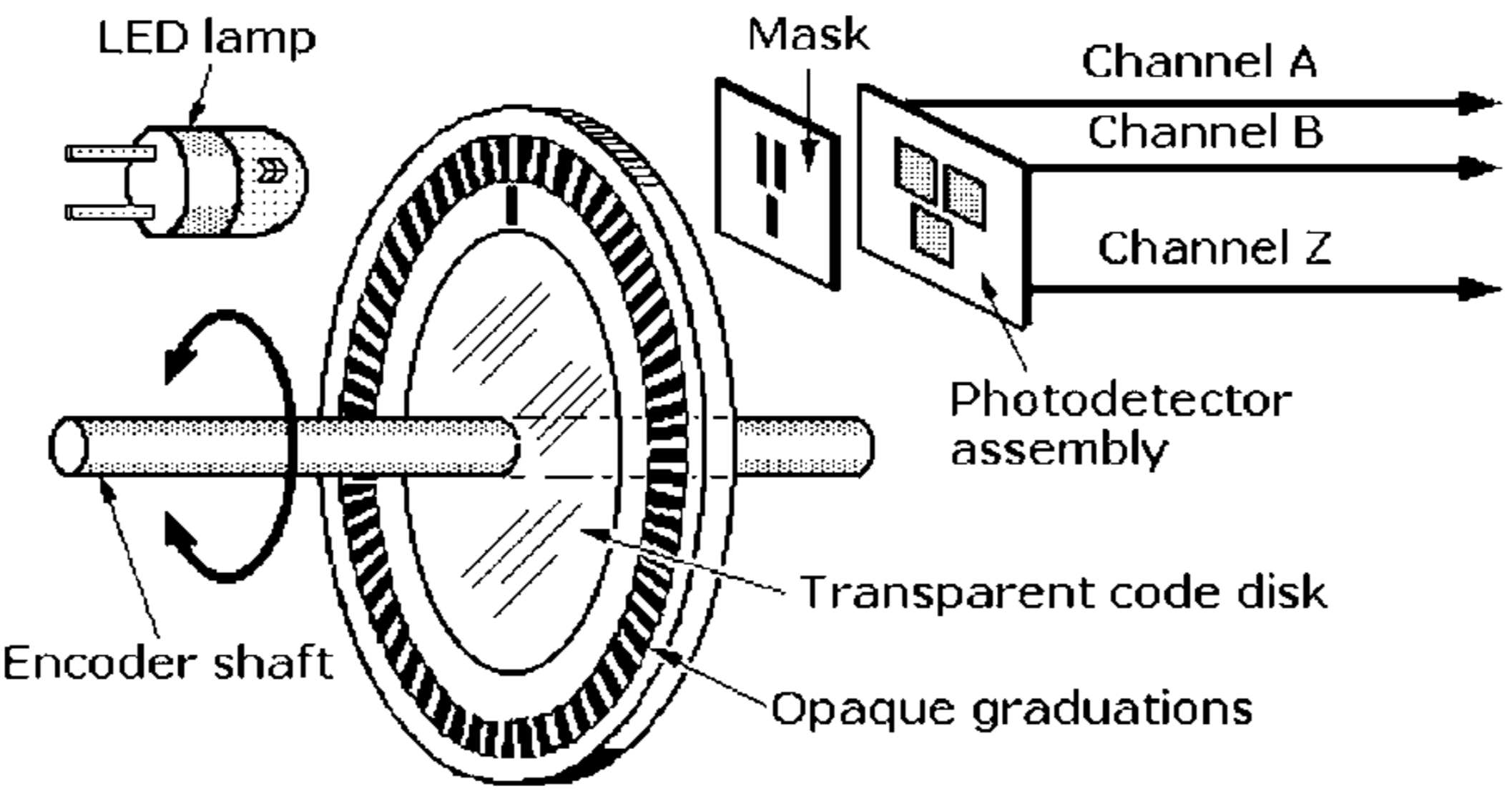}
\caption{\label{fig:encoder}Basic elements of an incremental optical rotary encoder. (adapted from~\cite{sclater2001mechanisms}).}
\end{figure}

\subsubsection{Optical Rotary Encoder}
\label{subsubsection:hw_encoder}
An optical rotary encoder with 1024~PPR\footnote{Pulse Per Revolution} was used to sense the actual positions of the servo as it moves to the commanded positions (Figure~\ref{fig:encoder}). The encoder reads the changes in angles (positions) as steps and those steps are later translated to positions (or change in positions). There are a variety of rotary encoders and the optical has higher precision~\cite{deshmukh2005microcontrollers, sclater2001mechanisms}. 
% The one used for this study offers 1024 PPR\footnote{Pulse Per Revolution}.

\subsection{Embedded System}
The first microcontroller was programmed to send the commanded servo positions to the servo and to maintain a uniform speed in servo motion to obtain sound readings from the rotary encoder. To keep the speed uniform, the servo positions were sent in steps of degrees to the servo in 232 milliseconds time units. When a position is successfully commanded to servo, the microcontroller sends a signal to the computer to affirm the completion of the command and the position commanded. Algorithm~\ref{alg:servo_code} is a snippet of the code.
The program embedded in the second microcontroller reads the steps recorded by the rotary encoder and sends it to the computer. When the first microcontroller reports the completion of the command of a position, the computer records the steps counted by the encoder's microcontroller with the commanded position reported by the servo's microcontroller. Algorithm~\ref{alg:encoder_code} is a snippet of the code.

% \begin{minipage} {0.46\textwidth}
% \fontsize{15pt}{30pt}\selectfont\bfseries
\begin{algorithm}
	% \setstretch{1.35}
	\centering
	\caption{Servo Controller}\label{alg:servo_code}
	\begin{algorithmic}[1]
	% \fontsize{9pt}{5pt}\selectfont\bfseries
	\Procedure{Servo\_Control}{}
	\State $\textit{degree} \gets 0$
	\State $\textit{position} \gets 0$
	\State $\textit{old\_position} \gets 0$
	\State $\textit{servo.move}(position)$
	\State $i \gets 0$
	\State $\textit{position} \gets \textit{Get\_Commanded\_Position}()$
	\State $\textit{degree} \gets \textit{old\_postions}-\textit{postion}$
	\BState \emph{loop}:
		\For{$\textit{i}\in \{0,\dots,\textit{absolute}(degree)$}
			\If {$\textit{degree}<0$}
				\State $\textit{servo.move}(old\_position+i)$
			\Else
				\State $\textit{servo.move}(old\_position-i)$
			\EndIf
			\State $\textit{delay}(232ms)$
		\EndFor
		\State $\textit{old\_position} \gets \textit{position}$
		\State $\textit{Report\_Complete\_Command}()$
		\State \textbf{goto} \emph{loop}.
	\EndProcedure
	\end{algorithmic}
\end{algorithm}

% \end{minipage}
% \hfill
% \begin{minipage}{0.46\textwidth}

\begin{algorithm}
	% \setstretch{1.35}
	\centering	
	\caption{Encoder Controller}\label{alg:encoder_code}
	\begin{algorithmic}[1]
	% \fontsize{9pt}{5pt}\selectfont\bfseries
	\Procedure{Encoder\_Control}{}
	\State $\textit{counter} \gets 0$
	\State $\textit{aState} \gets 0$
	\State $\textit{aLastState} \gets \textit{digitalRead}(Sensor\_A)$
	\BState \emph{loop}:
		\State $\textit{aState} \gets digitalRead(Sensor\_A)$
		\If {$\textit{aState} \ne \textit{aLastState}(j)$}
			\If {$\textit{digitalRead}(Sensor\_B) \ne \textit{Sensor\_A}$ }
				\State $\textit{counter} \gets \textit{counter} + 1$
			\Else
				\State $\textit{counter} \gets \textit{counter} - 1$
			\EndIf
			\State $\textit{Report}(counter)$
		\EndIf
		\State $\textit{aState} \gets \textit{aLastState}$
		\State \textbf{goto} \emph{loop}.
	\EndProcedure
	\end{algorithmic}
\end{algorithm}

% \end{minipage}

\subsection{Translating and Calibrating The Encoder's Signals}
The rotary encoder measures the rotation using optical signals captured as the encoder's shaft rotates~\cite{considine1957process}. These measurements are dependent on the rotation speed of the shaft. In order to maintain sound readings from the encoder, the speed of the servo is maintained uniform by controlling it from the servo's microcontroller. In addition, the actual readings of the encoder are steps (pulses) and not angles. Therefore, the readings are translated to angles by referring to the first round of logging. For example, hypothetically, if the first set of commanded positions were ${23,55,76}$ and its corresponding encoder reading were ${98,153,203}$, then these encoder readings are fixed for these angles and later encoder readings for positions changing rounds are compared to these encoder readings. Tests are performed for several servos, and deviation from these fixed encoder readings can be easily discovered, even for the case of a factory-faulty servo.

\section{Methodology}
\label{methodology}

The main concept 
underlying this experimental design was to determine the effect of time period on the reliability of a UAS aileron servo operated in various actuator positions continuously using real-time FDR data- derived positions. The objective of the research was to use a laboratory test-bed to assess the level of performance degradation and possible failure of a servo operated continuously at various actuated angles over a period of time.  
% of the testbed is a continuous operation of the servo in all positions collected from the real-time data provided by the FDR. The objective of the research was to use a laboratory test-bed to simulate the level of reliability of a servo that actuates the ailerons for small UAS.
The test-bed will be run at various commanded positions simulating real-life flight conditions and the actual output of the control surface actuated by the servo will be measured and compared. The set-up will be kept running over several cycles and time range to determine when the system will fail or become unreliable in terms of significant variations in output. The servo will be kept operating and the actual positions will be recorded for the commanded positions (Input commanded positions and actual position will be logged using a digital rotary encoder). 
It is hypothesized that with high fidelity and reliability of the conceptual design, the mean value of the encoder recorded position for the baseline period will not vary with increasing time and subsequent cycles. It is envisaged that the test-bed servo will continue to perform its intended function under stated operating conditions over a specified period of time without significant performance degradation. Significant variations in terms of the recorded values for the encoder recorded position compared to the commanded servo positions will suggest compromised reliability and probably failure of the system within time period. Thus, the hypothesises were as follows:
\begin{itemize}
	\item Null Hypothesis: $H_o : \mu_{baseline} = \mu_{period1} = \mu_{period2}$
	\item Alternative Hypothesis: $H_A : \mu_{baseline} \ne \mu{period 1} \ne \mu{period2}$
\end{itemize}

\subsection{Procedure}
\begin{figure}
\includegraphics[width=.98\textwidth]{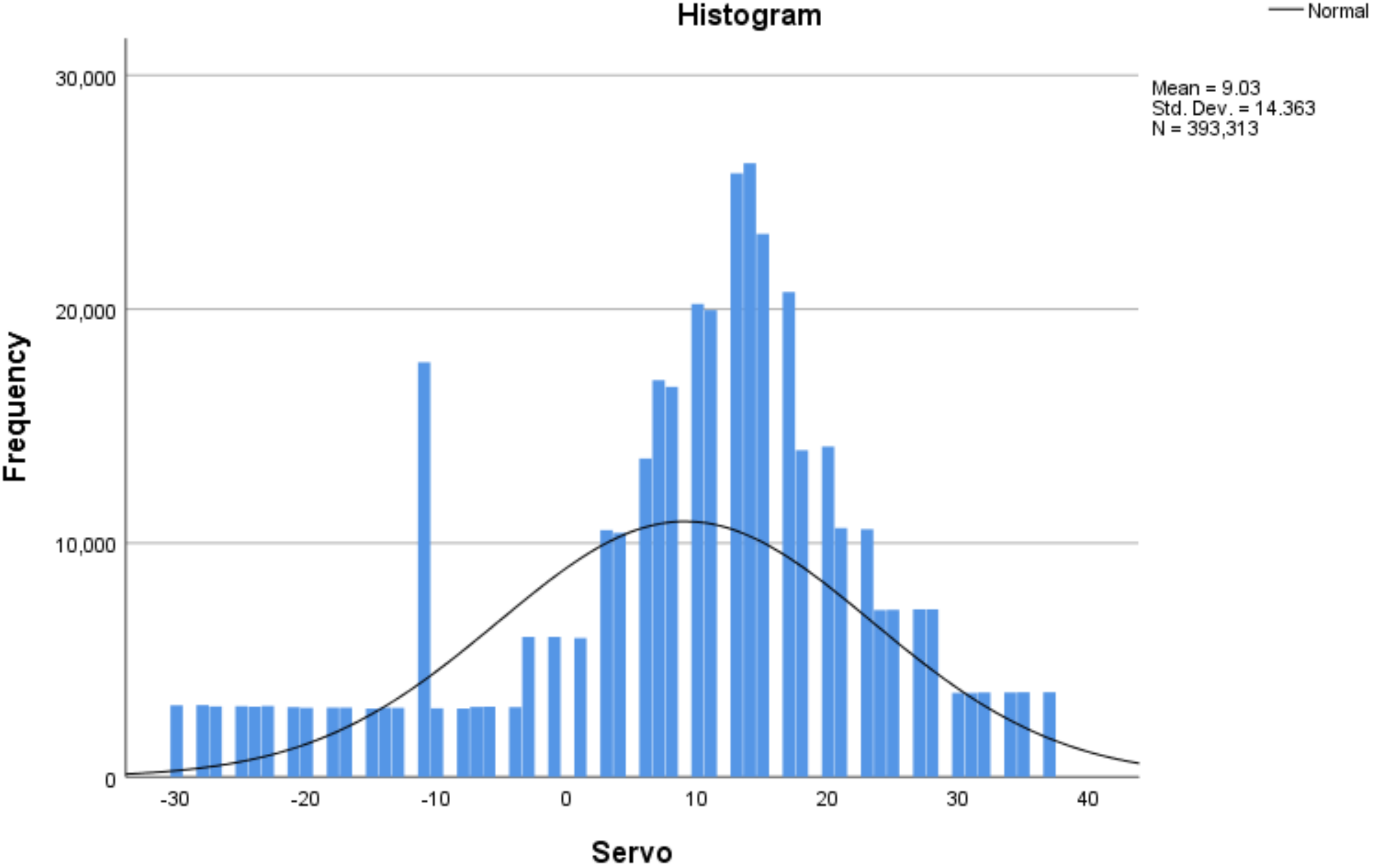}
\caption{\label{fig:histo_pos_postions}Histogram of commanded servo positions for entire dataset.}
\end{figure}

\begin{figure}
\includegraphics[width=.98\textwidth]{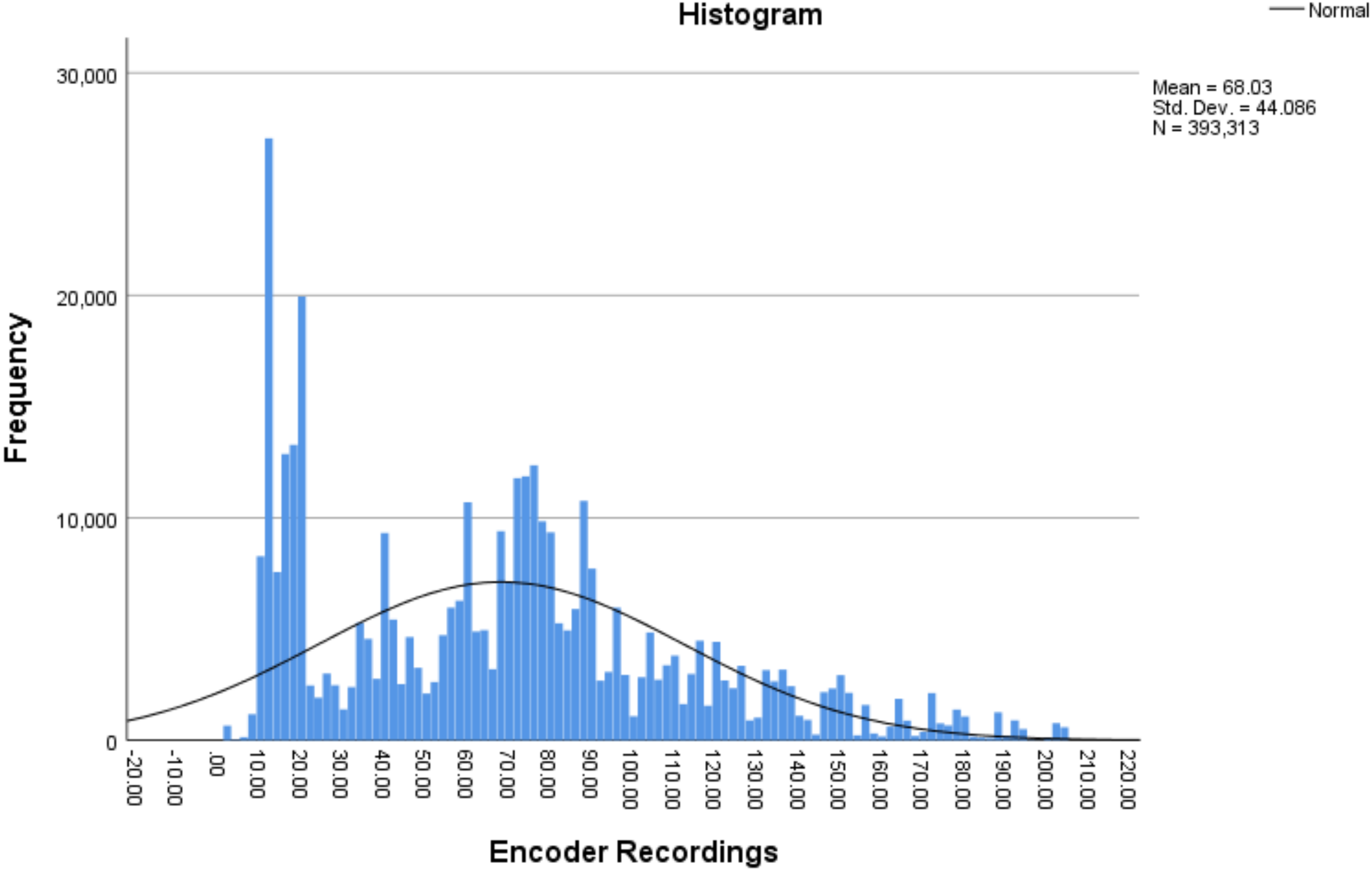}
\caption{\label{fig:histo_all_postions}Histogram of encoder recordings for entire dataset.}
\end{figure}

After almost 360 hours of continuous cycles (approximately 15 days), the data was extracted from a digital recorder which was part of the simulated set-up (an SD card). The data was recorded to a CSV file. A descriptive statistics summary showing the total cycles over the experimental periods for commanded positions of the servo and corresponding positions logged by optical encoder over time period are shown in Table~\ref{tab:descriptive_stats}. Figure~\ref{fig:histo_pos_postions} and Figure~\ref{fig:histo_all_postions} show the histogram with normal curve for the total data set of the servo position and encoder recordings respectively.

\begin{table}
	% \footnotesize
	\centering
  \caption{Descriptive Statistics of Study Variables Dataset}
  \label{tab:descriptive_stats}
  \begin{tabular}{lccccc}
	\multirow{2}{*}{} 
					& \multicolumn{1}{c}{\bfseries N} & \multicolumn{1}{c}{\bfseries Max} & \multicolumn{1}{c}{\bfseries Min} & \multicolumn{1}{c}{\bfseries Mean} & \multicolumn{1}{c}{\bfseries Std. Dev.} \\ 
					\cline{2-6}
				    & \multicolumn{1}{c}{Statistic} & \multicolumn{1}{c}{Statistic} & \multicolumn{1}{c}{Statistic} & \multicolumn{1}{c}{Statistic} & \multicolumn{1}{c}{Statistic} \\
    \toprule
	\makecell{Encoder \\ Recordings} 			& 393313	& .00	& 205.00	& 68.0345	& 44.08557 	\\
	\makecell{Commanded \\ Servo \\ Positions}	& 393313	& -30	& 37		& 8.22		& 14.601	\\
    \bottomrule
  \end{tabular}
\end{table}

The dataset of valid sample used for analysis for both variables ($N = 393313$) was further split into three periods based on cycles within a time frame (first $130,000$; second $130,000$; third $130,000$).  Out of each cache of data, random samples of 5000 were drawn using IBM SPSS\textsuperscript{\textregistered} 25 ``select dataset random'' function. The frequencies of various commanded positions of servo ( $-30$ to $+40$) were determined and the mean value of their output in the form of encoder recordings were also determined. The process was repeated for the other two datasets.

Three datasets, each with sample size ($n = 48$) were obtained and since they were derived from the same servo-unit of the test-bed, a one-way repeated-measure ANOVA design analysis was conducted to determine the levels of variability among the means of the outcome encoder readings for the three datasets. The objective was to determine if there were significant variations in encoder readings for commanded servo positions between the baseline (first $130,000$ cycles) and subsequent periods (second and third) till the end of experimental period and simulated failure of the servo on the designed test-bed. Table~\ref{tab:rand_sample_stats} shows the descriptive statistics of the  encoder readings for the random samples of commanded positions of the servo.

\begin{table}
	% \footnotesize
	\centering
  \caption{Descriptive Statistics of Study Variables Dataset}
  \label{tab:rand_sample_stats}
  \begin{tabular}{lccc}
	\multirow{1}{*}{}
					& \multicolumn{1}{c}{\bfseries Mean} & \multicolumn{1}{c}{\bfseries Std. Dev.} & \multicolumn{1}{c}{\bfseries N} \\
    \toprule
	$Cycle Period_1\_130$			& 54.3347	& 44.81194	& 48	\\
	$Cycle Period_2\_260$			& 59.9299	& 50.30706	& 48	\\
	$Cycle Period_3\_390$			& 66.6863	& 55.13150	& 48	\\
    \bottomrule
  \end{tabular}
\end{table}

\section{Results and Discussions}
\label{results}

The study sought to evaluate the reliability of a designed-test bed simulating a UAS aileron servo-unit by using an optical encoder recorder to log the actual position of the test-unit derived from an input command servo position. The study hypothesized that there will be no significant differences in the mean positions recorded by the optical encoder for the entire spectrum of commanded servo positions within time period [baseline ($130,000$ cycles), period two ($260,000$ cycles) and period three ($390,000$ cycles)]. 

The rationale for the study hypothesis was that if the designed test-bed has high reliability then operational cycles within the time period simulated will not significantly affect the accuracy of the input commanded position and the actual position servo position recorded by the optical encoder. The alternate hypothesis was that there will be significant differences in the mean of the outputs over periods as unit approaches imminent failure due to continuous operations over time period.

A one-way repeated measures ANOVA was performed. The mean values of encoder recordings of input commands for the servo positions ( -30 to +40) was derived from a random sample of 5000 drawn from the three periods. The derived dataset ($N=48$) for each three periods were used for the analysis. The sample size was large enough to for the assumptions of multivariate normality to be satisfied. 
The Mauchly test was performed to assess possible violation of the sphericity assumption; this was significant: Mauchly’s $W = .523$, $\chi^2(2) = 29.82$, $p = 0.000$; suggesting a possible violation of sphericity assumption~\cite{field2017discovering}. The Greenhouse-Geisser $\epsilon$ value of $0.677$ was not close to $1.00$, and correction was made to the degrees of freedom used to evaluate the significance of the F ratio. The overall F for differences in mean encoder recordings across the three cyclic periods was statistically significant: $F(1.35, 63.64) = 21.02$, $p = 0.000$; the corresponding effect size was a partial $\eta^2$ of $0.369$.
In other words, after stable individual differences in various commanded servo positions are considered, about $37\%$ of the variance in encoder position recorded was related to time. Note that with the $epsilon$ correction factor was applied to the degrees of freedom for $F$, the obtained value of F for differences in mean encoder recording among levels of commanded servo positions remained statistically significant.

Planned contrasts were obtained to compare mean encoder recorded values for each of the two time periods with the mean encoder recording during baseline [$(M_{baseline} = 54.33; SE =6.468; \quad 95\%CI,\quad 41.32 – 67.35)$].
Mean encoder recordings during the second period [$(M_{2nd\_period} = 59.93; \quad SE = 7.261; \quad 95\%CI, 45.32-74.54)$] was significantly higher than baseline encoder recorded value $F(1, 47) = 27.49$, $p = 0.000$. Mean encoder recorded value for the third period $(M_{3rd\_period} = 66.69; SE =7.958; 95\%CI, 50.68-82.70)])$ was significantly higher than the baseline and second period $F(1, 47) = 9.79$, $p = .003$.
Due to the possible violations of sphericity and corrections for the $F$ using Greenhouse-Geisser estimate of sphericity, a further multivariate analysis was conducted, and all three test statistics were significant (using $\alpha = .05$ as the criterion). The encoder recorded values for commanded servo positions were significantly affected by time, Pillai’s Trace $V = .479$, $F(2, 46) = 21.15$, $p = 0.000$. The corresponding $\eta^2$ effect size of $0.48$ indicated a strong effect.

\begin{figure}
\includegraphics[width=.98\textwidth]{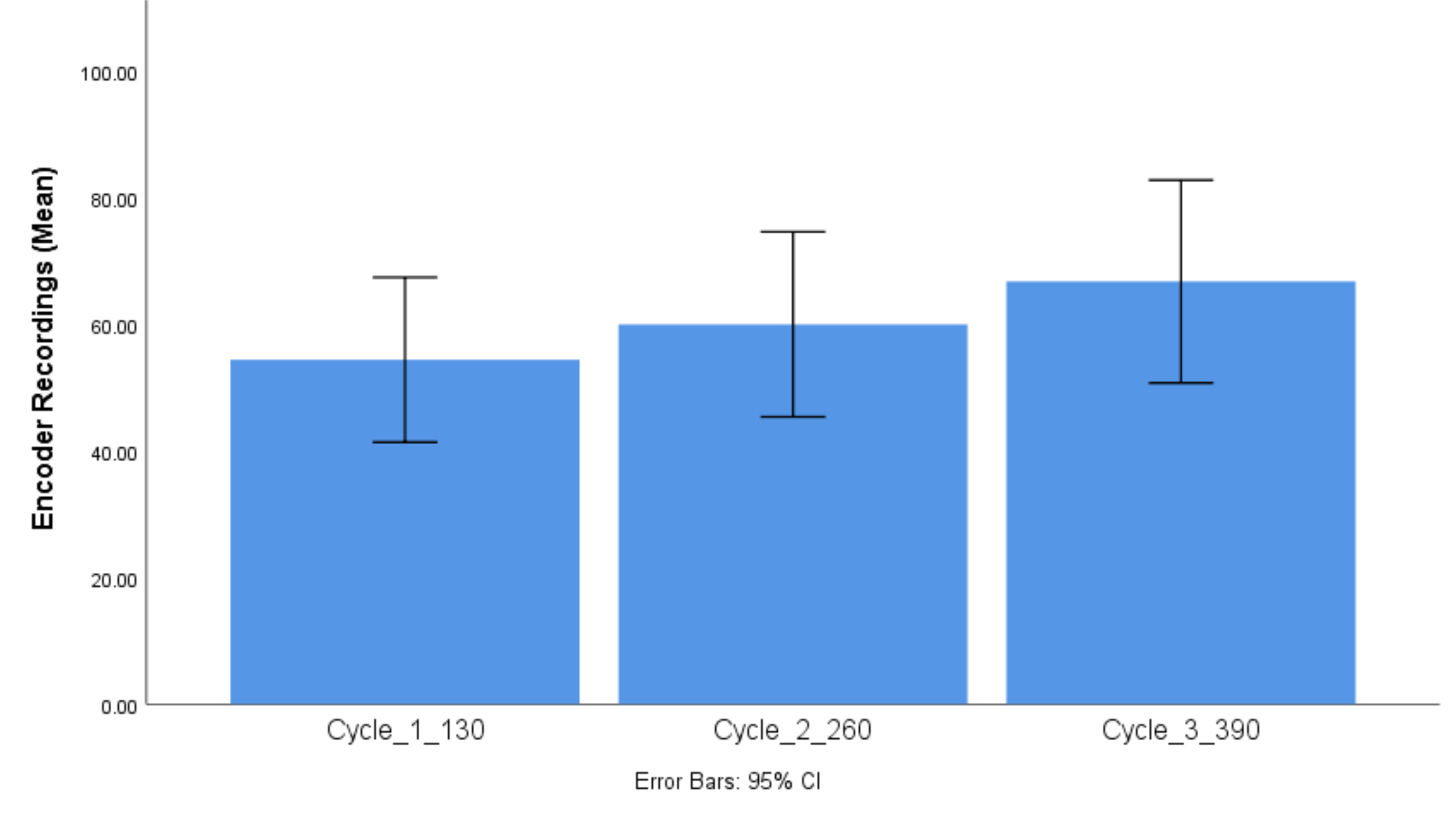}
\caption{\label{fig:barplot_mean_encoder}Simple bar ( mean) of encoder recordings at various period cycles.}
\end{figure}

\begin{figure}
\includegraphics[width=.98\textwidth]{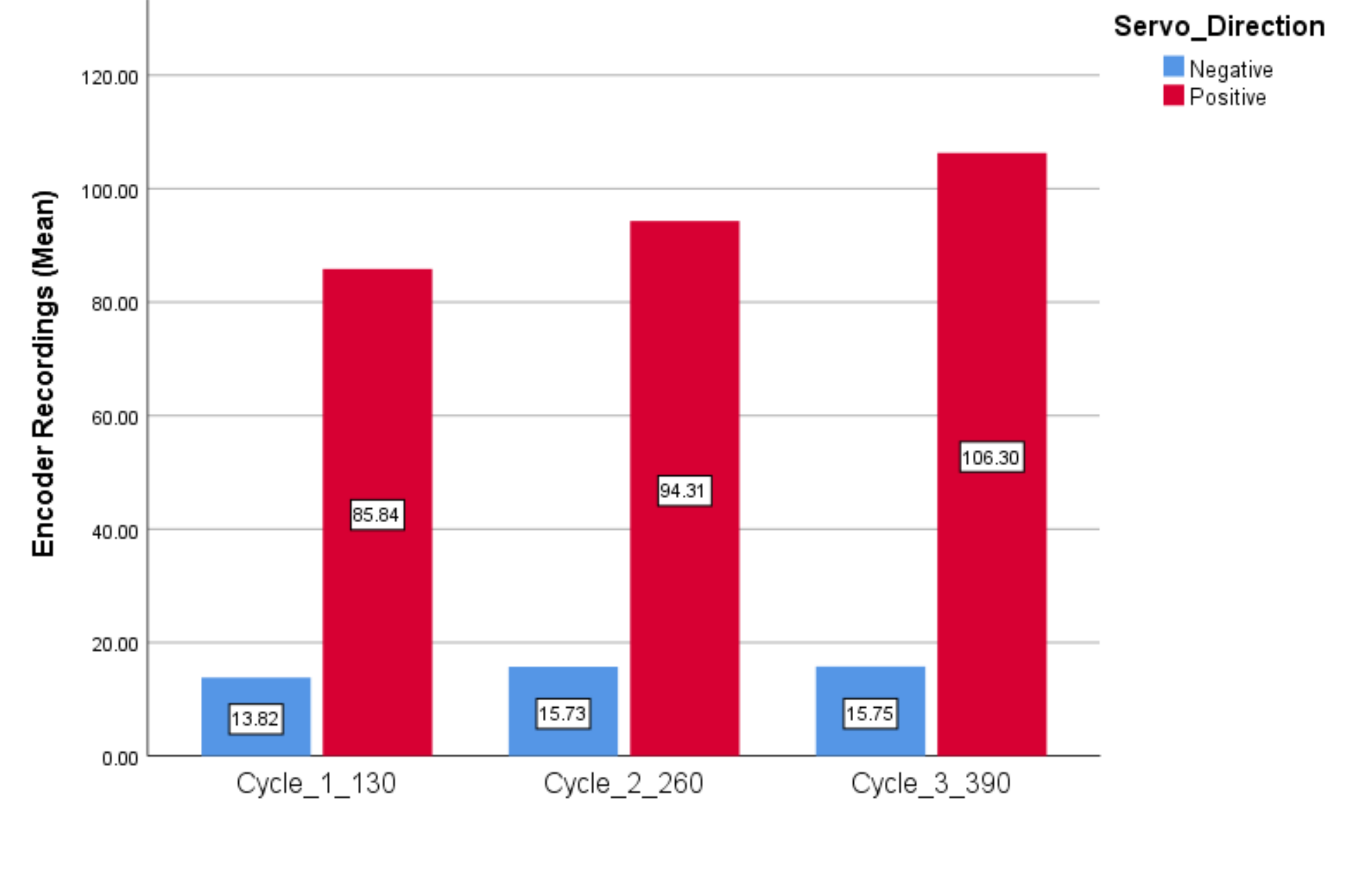}
\caption{\label{fig:barplot_clustered}Clustered bar (mean) of encoder recordings with servo directions.}
\end{figure}

\begin{figure}
\includegraphics[width=.98\textwidth]{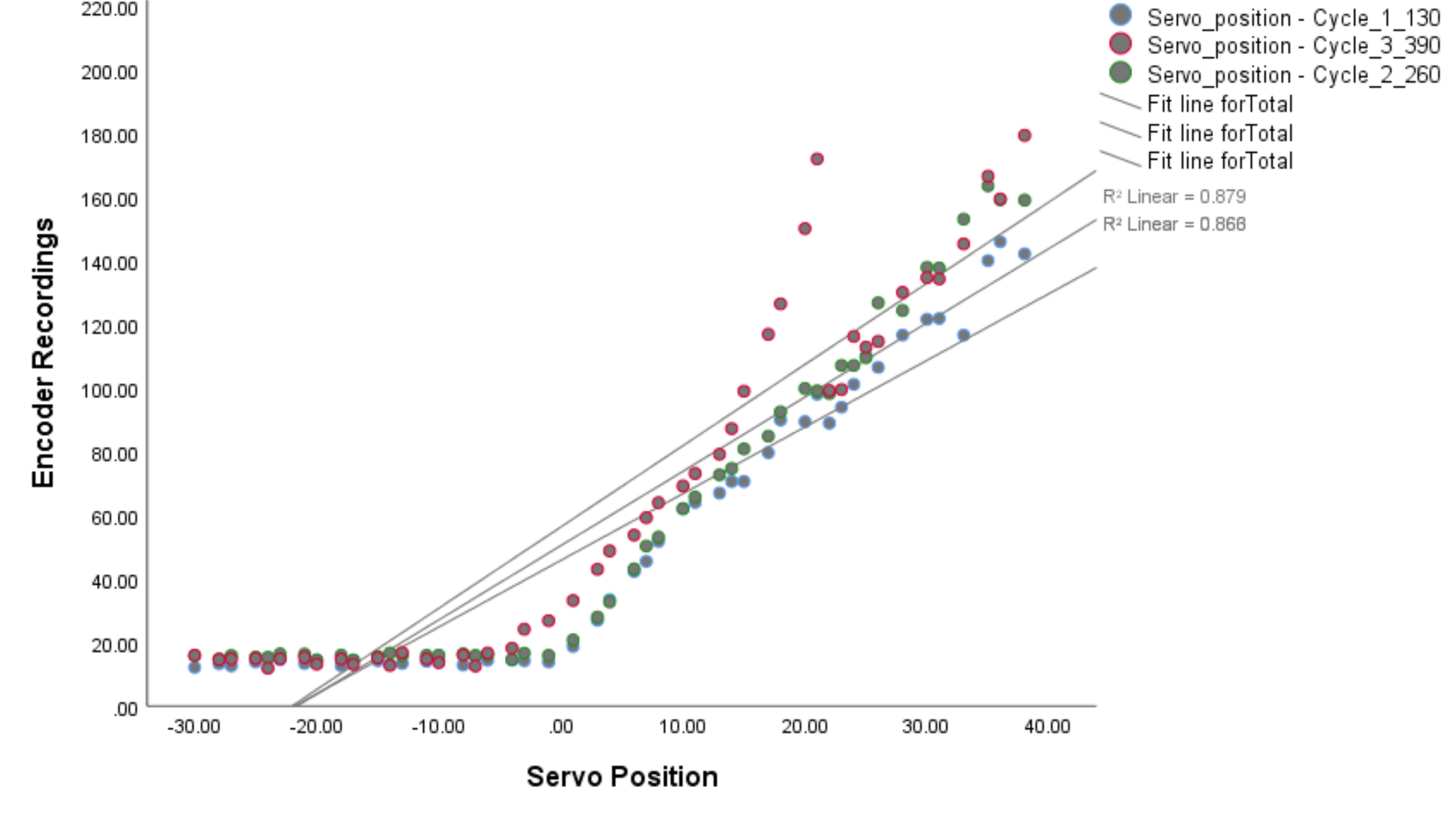}
\caption{\label{fig:results_plot}Simple scatter with fit line of encoder recordings against commanded servo positions.}
\end{figure}

To highlight the variations graphically,
Figure~\ref{fig:barplot_mean_encoder}
shows a simple bar (mean) of encoder recordings at various period (cycles) for commanded servo positions.
Figure~\ref{fig:barplot_clustered}
depicts a clustered bar (mean) of encoder recordings with servo directions ( negative and positive) and
Figure~\ref{fig:results_plot}
shows a simple scatter with fit line of encoder recordings against commanded servo positions. The scatter plot suggests a significant variation in encoder recording during the third period especially for commanded servo position range of $+10$ to $+30$. The scatter also suggests that there are minimal variations in the negative commanded positions of the servo across all spans of the periods studied as shown by
Figure~\ref{fig:barplot_clustered}. Though this was not anticipated, the reason behind this could be that the used servo in the experiment was faulty hardware-wise (a manufacturing fault). Another explanation might be that the negative aileron angles do not have high values and are not frequent compared to the positive ones (Figure~\ref{fig:histo_pos_postions}).

% \TODO{Discussions:: How does our results compare or contrast to earlier findings from extant research on reliability of components with time in aviation? Use the literature review sources to make the analysis and comparison for the discussions? Why were there minimal variations at the negative commanded inputs and outputs over time but significant variations with the positive? Why such variations especially from +1-+30? What are the operational, policy and research implications of the study?}

Comparing the current findings with that of Long~\etal, reliability analysis based on MBTF or periodic analysis should be a focus not only in the more formal military UAS applications, but the varied and more common COTS manufactured components. When reliability becomes a focus of management, it improves markedly, even for systems developed and transitioned under accelerated circumstances when quality and reliability have to take second place behind other priorities. With more studies and research data on the current focus of study, it is envisaged that COTS manufactured  UAS component’s  reliability will improve over time. This is because such studies will produce  knowledge and experience with the technology gained both in the production plant and in the field by users.  This is very important in building a veritable base of the body of knowledge~\cite{anderson1976reliability}.

This study underpins Casewell \& Dodd's earlier findings~\cite{caswell2014improving} that commercial grade UAS components are more prone to failures than military-grade components. It is suggested that non-sensitive and non-classified data on reliability of UAS components in military application be shared with academia and civil industry players so that technological synergy and spin-offs can improve COTS UAS component reliability. Even though this study did not operate the test-bed to attrition, the performance discrepancy (reliability) over time buttresses the need for more research into that area of study. This study also adds a quantitative and statistical dimensions to earlier findings of Uhlig~\cite{neogi2007engineering}. who studied COTS as a cause of failure in UAS systems. Their study investigated the integration of the specific sub-systems and added some redundancy to the components used to see its influence on the reliability of the UAS. It is hopes that time to failure will become an essential metric for COTS standards as suggested in this study.

\section{Conclusion And Future Work}
\label{sec:conclusion}

The researchers assessed the reliability of a laboratory designed UAS component test-bed operated using real-world data collected from a Boeing Scan Eagle\textsuperscript{\textregistered} UAS aileron servo unit via a flight data recorder. The study hypothesized that a test-bed unit replicating a UAS aileron servo motor’s reliability in terms of a base-line measured encoder output of commanded servo position’s will not be significantly different after double and triple periods of cycle time. The discrepancy between servo-commanded positions and actual positions were recorded using an optical digital encoder. A repeated-methods ANOVA was used to determine if there existed significantly in the mean values of the commanded output positions in the three periods. Results suggest significant relationship between testbed reliability and cycle periods, with lower reliability as time increases.

% The servo motor test-bed introduced in this study addresses one of the persisting problems of using COTS parts on UAVs without actually having reliability studies for them.
The test-bed considered an actual case study about a servo that is serving in one of the most successful long-endurance UAVs. The forces acting on the servo were simulated using a linearly-increasing-force extension mechanical springs. The hardware structures were CAD designed, taking into consideration the freedom of movement to align the servo shaft with the rotary encoder shaft, to alleviate any additional forces. 
% other than the ones actually in the problem subject of the design.

Later, the code used to control the system and collect the data was developed, deployed, and tested. Finally, an experiment was conducted to illustrate the outcome of the proposed system. The results showed interesting variability in the servo performance as it operates through time. 
Though the servo did not completely fail, a slight difference in the commanded position can result in a devastating situation as a UAS maneuvers.

For future work, some hardware can be enhanced. The apparatus used copper wires to attach the servo to the springs, which is not ideal as copper might be more malleable than desired causing extensions in the attaching links and that would affect the designed acting force of the muscle springs. Therefore, a stiffer material like stainless steel wires or aluminum thin rods would be consider as a replacement. Also, the fixation of the servo base can be better designed. 

Further, additional experiments shall be carried out on a collection of servos of the type used on the UAS subject of the introduced case-study, which shall fill an essential gap in extant literature on UAS reliability studies and also provide data-driven benchmarks for aviation regulators during UAS components certification processes. 

% The findings from such future studies will also enhance the overall aim of continuous monitoring and improvements in UAS safety globally.

% This shall add to the findings of this study and offer an empirical statistical model for the actual servos used on operating UAS.

Ultimately the outcome of this future work would contribute the confidence in BVLOS of UAS and would induce the regulatory forms to commercially permit these kind of flights. 
% to offer useful information for the operators of one of the most successful UAVs in unmanned aviation.
% This shall be used for a statistical study to enhance and support the findings, which might offer useful information for the operators of one of the most successful UAVs in unmanned aviation.

\bibliographystyle{unsrt}  
\bibliography{biblography.bib}

\end{document}